\documentclass[sigconf]{acmart}
\AtBeginDocument{%
  }

\copyrightyear{2026}
\acmYear{2026}
\setcopyright{cc}
\setcctype{by}
\acmConference[WWW '26]{Proceedings of the ACM Web Conference 2026}{April 13--17, 2026}{Dubai, United Arab Emirates}
\acmBooktitle{Proceedings of the ACM Web Conference 2026 (WWW '26), April 13--17, 2026, Dubai, United Arab Emirates}
\acmPrice{}
\acmDOI{10.1145/3774904.3793027}
\acmISBN{979-8-4007-2307-0/2026/04}

\usepackage{subcaption}
\usepackage{multirow}
\usepackage{pifont}
\usepackage{xcolor}
\usepackage{colortbl}
\usepackage{booktabs}
\usepackage{tabularx}
\usepackage{threeparttable}
\usepackage{enumitem}
\usepackage[most]{tcolorbox}
\newcommand{\blue}[1]{\textcolor{black}{#1}}

\definecolor{comment}{RGB}{200, 160, 0}
\newcommand{\yifan}[1]{\textcolor{black}{#1}}

\definecolor{myblue}{HTML}{14517c} 
\begin{document}

\title{Mind the Ambiguity: Aleatoric Uncertainty Quantification in LLMs for Safe Medical Question Answering}
\author{Yaokun Liu}
\affiliation{%
  \institution{University of Illinois Urbana-Champaign}
  \city{Champaign}
  \state{IL}
  \country{United State}
}
\email{yaokunl2@illinois.edu}

\author{Yifan Liu}
\affiliation{%
  \institution{University of Illinois Urbana-Champaign}
  \city{Champaign}
  \state{IL}
  \country{United States}
}
\email{yifan40@illinois.edu}

\author{Phoebe Mbuvi}
\affiliation{%
  \institution{University of Illinois Urbana-Champaign}
  \city{Champaign}
  \state{IL}
  \country{United States}
}
\email{pmbuvi2@illinois.edu}

\author{Zelin Li}
\affiliation{%
  \institution{University of Illinois Urbana-Champaign}
  \city{Champaign}
  \state{IL}
  \country{United States}
}
\email{zelin3@illinois.edu}

\author{Ruichen Yao}
\affiliation{%
  \institution{University of Illinois Urbana-Champaign}
  \city{Champaign}
  \state{IL}
  \country{United States}
}
\email{ryao8@illinois.edu}

\author{Gawon Lim}
\affiliation{%
  \institution{University of Illinois Urbana-Champaign}
  \city{Champaign}
  \state{IL}
  \country{United States}
}
\email{gawonl2@illinois.edu}

\author{Dong Wang}
\affiliation{%
  \institution{University of Illinois Urbana-Champaign}
  \city{Champaign}
  \state{IL}
  \country{United States}
}
\email{dwang24@illinois.edu}

\renewcommand{\shortauthors}{Yaokun Liu et al.}




\begin{abstract}

The deployment of Large Language Models in Medical Question Answering is severely hampered by ambiguous user queries, a significant safety risk that demonstrably reduces answer accuracy in high-stakes healthcare settings.
In this paper, we formalize this challenge by linking input ambiguity to aleatoric uncertainty (AU), which is the irreducible uncertainty arising from underspecified input. 
To facilitate research in this direction, we construct CV-MedBench, the first benchmark designed for studying input ambiguity in Medical QA.
Using this benchmark, we analyze AU from a representation engineering perspective, revealing that AU is linearly encoded in LLM's internal activation patterns.
Leveraging this insight, we introduce a novel AU-guided “Clarify-Before-Answer” framework, which incorporates AU-Probe—a lightweight module that detects input ambiguity directly from hidden states.
Unlike existing uncertainty estimation methods, AU-Probe requires neither LLM fine-tuning nor multiple forward passes, enabling an efficient mechanism to proactively request user clarification and significantly enhance safety.
Extensive experiments across four open LLMs demonstrate the effectiveness of our QA framework, with an average accuracy improvement of $9.48\%$ over baselines. Our framework provides an efficient and robust solution for safe Medical QA, strengthening the reliability of health-related applications. 
The code is available at \href{https://github.com/yaokunliu/AU-Med.git}{https://github.com/yaokunliu/AU-Med.git}, and the CV-MedBench dataset is released on Hugging Face at \href{https://huggingface.co/datasets/yaokunl/CV-MedBench}{https://huggingface.co/datasets/yaokunl/CV-MedBench}.


\end{abstract}

\maketitle

\section{Introduction}\label{sec:intro}

\begin{figure}[t]
    \centering
    \includegraphics[width=\linewidth]{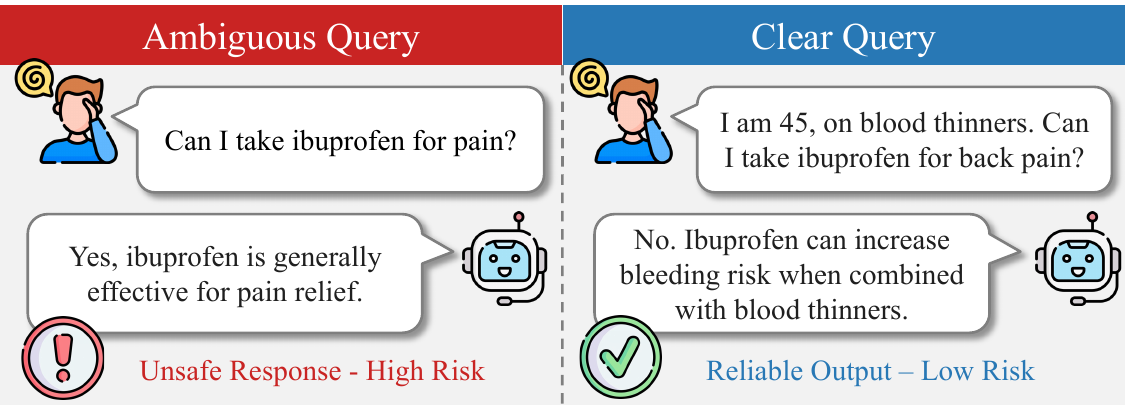}
    \vspace{-0.5cm}
    \caption{Effect of input ambiguity on medical LLM responses. Ambiguous queries lead to unsafe outputs, while clear queries enable reliable reasoning.}
    \label{fig:intro_ambiguity_demo}
\vspace{-3mm}
\end{figure}

\begin{figure}[t]
    \centering
    \includegraphics[width=1\linewidth]{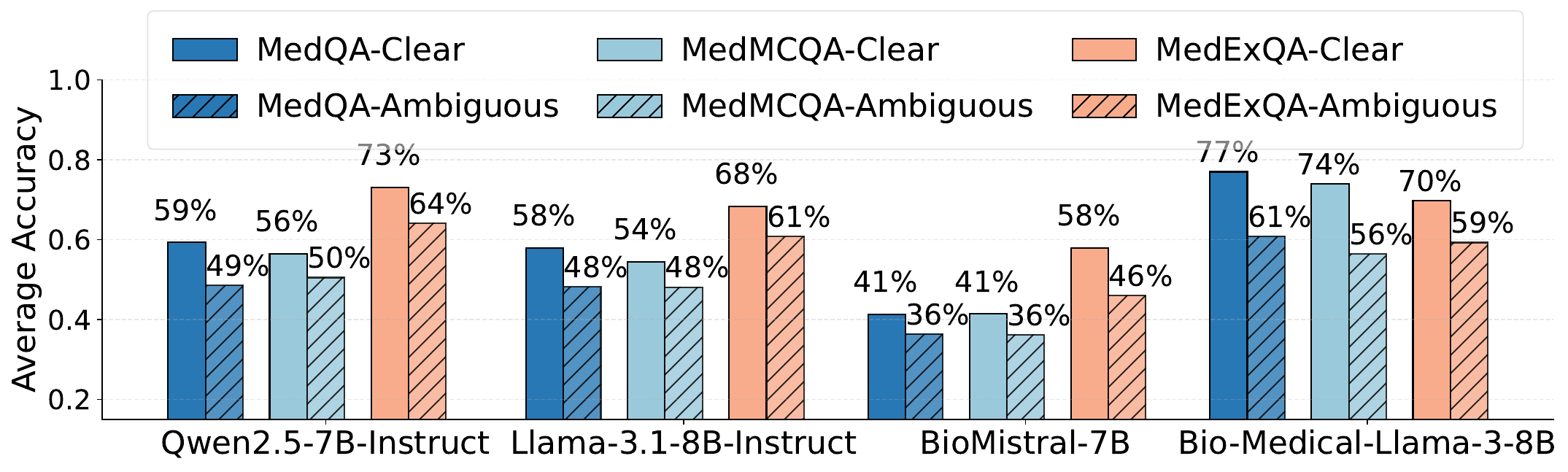}
    \vspace{-0.5cm}
    \caption{
    Input ambiguity reduces answer accuracy across models and datasets.
    }
    \label{fig:accuracy_ambiguity}
\vspace{-5mm}
\end{figure}

Large Language Models (LLMs) are increasingly adopted as convenient and affordable tools for medical question answering (Medical QA)~\cite{wang2024large,sellergren2025medgemma,wu2024pmc}. 
In 2024, around 31\% of U.S. adults reported using LLMs to seek preliminary explanations or advice about symptoms rather than immediately consulting healthcare professionals~\cite{choy2024can}.
Although LLMs broaden access to medical knowledge, they are also exposed to a critical safety issue: users often submit highly ambiguous queries~\cite{atf2025challenge}. For example, a user might ask \textsf{“Can I take ibuprofen for pain?”} without mentioning essential demographics or symptom details. 
\blue{This input ambiguity prevents LLM from identifying patient-specific contraindications and raises the risk of unsafe medical advice~\cite{wang2024safety}, as shown in Figure~\ref{fig:intro_ambiguity_demo}. 
Our empirical evidence further confirms this safety concern: we observe a significant drop in response accuracy across LLMs when queries are ambiguous (Figure~\ref{fig:accuracy_ambiguity}).}
In high-stakes medical decision support, these unsafe responses are consequential, as they may mislead users and endanger patient health. Therefore, there is an urgent need for methods to detect and manage the risks posed by ambiguous queries.

A promising solution to detect ambiguity risks lies in Aleatoric Uncertainty (AU) quantification, which captures the \yifan{ uncertainty inherently in the input. Specifically,} 
AU represents the intrinsic variability in the underlying answer distribution, which constitutes an important component of model uncertainty~\cite{hullermeier2021aleatoric}. In Medical QA, aleatoric uncertainty directly reflects the degree of ambiguity in user queries~\cite{se,min2020ambigqa}. For example, when a user asks \textsf{"Can I take ibuprofen to relieve pain?"} without specifying the type of pain or clinical context, the model may generate a variety of possible answers, resulting in a higher AU. \blue{In this work, we aim to quantify AU for user queries and leverage it as a principled indicator for identifying inputs whose ambiguity poses a risk to answer safety.}

\blue{However, existing methods for uncertainty quantification (UQ) in LLMs face two major challenges when applied to AU quantification in medical QA.} 
Firstly, many UQ methods estimate total predictive uncertainty rather than isolating the aleatoric component~\cite{entropy,sar,ptrue,gnll}. These methods are validated on controlled benchmarks where input ambiguity (i.e., AU) is minimal or absent. While these methods perform well under such idealized conditions, prior studies reveal that their performance degrades sharply when confronted with real-world query ambiguity~\cite{tomov2025illusion}. 
\yifan{Secondly, existing methods that explicitly quantify AU in LLMs are computationally prohibitive, making them unsuitable for time-sensitive clinical scenarios where response latency is critical. }Specifically, existing methods can be categorized into sampling-based~\cite{hou2023decomposing,gao2024spuq,ling2024uncertainty,cole2023selectively} and ensemble-based methods~\cite{DBLP:conf/icml/AhdritzQ0BE24, fort2019deep, he2020bayesian}.
Sampling-based methods require multiple rounds of generation to capture variability across \blue{outputs}, 
\blue{while ensemble-based methods compare predictions across several independently trained QA models.}
\yifan{Both categories} require repeated forward passes or model replicas, making them computationally expensive and impractical for deployment.
\blue{These limitations motivate our key research question: \textit{How can the AU of medical QA be quantified both accurately and efficiently to proactively detect ambiguity risks?}}


Recent advances in representation engineering show that LLMs encode high-level semantic attributes, such as sentiment and factuality, along approximately linear directions in their activation space, \yifan{also referred to as the Linear Representation Hypothesis}~\cite{park2023linear, zou2022representation, rimsky2024steering}.
Inspired by these findings, in this paper, we examine whether AU is also linearly encoded in the internal states of LLMs. Our analysis reveals a striking pattern: activations from low-AU (clear) and high-AU (ambiguous) medical queries can be separated by a \textit{linear} boundary. This observation indicates that AU can be estimated directly from internal activations of LLMs, allowing inference-time AU quantification before answer generation.

\blue{Building on this observation, we propose the AU-Guided “Clarify-Before-Answer” Medical QA Framework, which detects AU directly from the internal states of LLMs before answer generation, enabling timely detection and proactive mitigation of ambiguity-induced risks. This framework is enabled by three core contributions:
\begin{itemize}[leftmargin=*]
\item \textbf{Clear-to-Vague Benchmark (CV-MedBench).}
We construct CV-MedBench, the first Medical QA benchmark that includes systematically generated ambiguous questions, addressing the gap of existing datasets that do not capture realistic user input ambiguity. This benchmark undergoes rigorous human verification, validating its quality and suitability for the systematic studies of input ambiguity in Medical QA.
\item \textbf{Aleatoric Uncertainty Probe (AU-Probe).}
We introduce the AU-Probe, a lightweight plug-in module designed to quantify AU directly from the LLM's internal activations. By operating on hidden states, the AU-Probe requires no \yifan{repeated forward passes} or LLM fine-tuning, ensuring real-time and easily deployable AU estimation before model answer generation.
\item \textbf{AU-Guided “Clarify-Before-Answer” Pipeline.} 
We propose the first framework that employs AU as an explicit guiding signal for a "Clarify-Before-Answer" pipeline. In the first stage, AU-Probe estimates the AU score of each incoming question before answer generation, providing an early indicator of input ambiguity. When the AU score exceeds the safety threshold, the system proactively requests clarification from the user. Once the query is sufficiently clarified, in the second stage, the LLM proceeds with answer generation. This pipeline enables early risk detection, reduces ambiguity-induced errors, and substantially enhances the safety and reliability of Medical QA systems.
\end{itemize}}
Extensive experiments across four open-source LLMs demonstrate the effectiveness of our methods: AU-Probe achieves substantially stronger ambiguity discrimination with an average AUROC improvement of 49.24\% over baselines, while the AU-guided clarification pipeline improves answer accuracy of LLMs by 9.48\% with high efficiency. 
\blue{Overall, our framework strengthens the safety and reliability of Medical QA systems, supporting more trustworthy health-related applications and clinical decision support.}



\section{Related Works}

\subsection{LLMs for Medical QA}

Medical QA is a particularly demanding task that requires precise domain-specific reasoning under strict accuracy requirements, as even minor errors in clinical scenarios can lead to unsafe or misleading guidance, especially for non-expert users. To address these challenges, recent efforts have applied general-purpose and domain-adapted LLMs to medical QA~\cite{xu2025survey, thirunavukarasu2023large}. Med-PaLM~\cite{singhal2023large} adapts instruction-tuned models to medical domains and achieves expert-level accuracy on benchmark exams. PMC-LLaMA~\cite{wu2024pmc} leverages PubMed Central articles to enhance domain alignment, while BioGPT~\cite{luo2023biomedgpt} pretrains on biomedical corpora for improved biomedical reasoning. MedGemma~\cite{sellergren2025medgemma} \yifan{optimizes a general-purpose LLM for multimodal medical text and image comprehension.}

While these models focus on enhancing knowledge coverage and generation accuracy, little attention has been paid to input ambiguity--a common source of risk in real-world applications, leaving a critical gap in the safe and robust deployment of LLM-based medical QA systems. \yifan{Our method directly bridges this gap by enabling early detection of input ambiguity.}

\vspace{-0.1cm}
\subsection{Uncertainty Quantification in LLMs}\label{sec:uq}

Uncertainty quantification aims to assess the reliability of model outputs and has become an essential component for detecting hallucinations and improving the trustworthiness of LLMs. The total uncertainty of LLMs can be decomposed into epistemic uncertainty (EU) and aleatoric uncertainty (AU). A variety of UQ methods have been explored to quantify total uncertainty, including log-probability entropy~\cite{entropy,takayama2019relevant,van2022mutual}, self-consistency across sampled generations~\cite{se,sar,liu2025reasoning}, LLM-as-a-judge~\cite{ptrue, tian2023just}, and internal representation–based estimators~\cite{kossen2024semantic,vazhentsev2025uncertainty}. While effective when inputs are clear, these methods for total uncertainty estimation have been shown to degrade significantly when queries contain ambiguity, revealing their inability to effectively model aleatoric uncertainty arising from vague user inputs~\cite{tomov2025illusion}.
To explicitly quantify aleatoric uncertainty, recent studies have proposed sampling-based~\cite{hou2023decomposing,gao2024spuq,ling2024uncertainty,cole2023selectively} and ensemble-based~\cite{DBLP:conf/icml/AhdritzQ0BE24, fort2019deep, he2020bayesian} strategies. Sampling-based methods estimate aleatoric uncertainty by generating multiple clarifications or perturbations of the same input to measure output variability, while ensemble-based approaches aggregate predictions across auxiliary LLMs to isolateinput-driven uncertainty from epistemic uncertainty. However, both categories require multiple forward passes or additional model replicas, introducing substantial computational cost and latency that make them unsuitable for real-time medical QA systems.

These limitations highlight the need for an efficient aleatoric uncertainty quantification method that can operate within the response-time constraints of medical QA.

\vspace{-0.1cm}
\subsection{Representation Engineering}
Recent advances in representation engineering reveal that LLMs encode high-level semantic and behavioral properties along approximately linear directions in the hidden activations, enabling detection and manipulation of complex attributes through simple linear operations~\cite{zou2022representation, park2023linear, rimsky2024steering}. 
Prior studies have demonstrated this for diverse attributes: Tigges et al.~\cite{tigges2023linear} show that sentiment is captured by a single direction in the residual stream, with ablation degrading prediction accuracy; Waldis et al.~\cite{waldis2025aligned} show that toxic vs. non-toxic content is linearly separable in hidden activations, enabling detoxification through vector-based steering; 
and Arditi et al.~\cite{arditi2024refusal} demonstrate that safety-aligned refusal behavior is governed by a dominant direction, allowing precise behavioral control.

Despite this progress, the linear representational structure of aleatoric uncertainty, which arises from input ambiguity, remains unexplored. Building on these insights, we \blue{present the first analysis examining} whether representations of clear versus ambiguous medical questions form linearly separable patterns in hidden activations. Our analyses confirm such geometry, motivating a lightweight probe for efficient AU quantification directly from internal states.
\section{\blue{Clear-to-Vague Medical QA Benchmark}}\label{benchmark}

\blue{Existing public medical QA datasets predominantly contain questions written in clear and detailed clinical language.   
While suitable for evaluating diagnostic reasoning, such questions do not reflect how ordinary users typically express medical concerns in real-world settings, where queries are often vague, incomplete, or semantically imprecise. To address this gap and enable the study of aleatoric uncertainty in medical QA, we construct the \textbf{Clear-to-Vague Medical QA Benchmark} (\yifan{CV-MedBench}), which pairs existing clear medical query resources with their ambiguous counterparts.}

In particular, we collect clear questions from three public medical QA datasets MedQA~\cite{medqa}, MedMCQA~\cite{medmcqa}, and MedExQA~\cite{medexqa}, and prompt a LLM 
(e.g., GPT-4o) to rewrite each clear question $\mathbf{x}_{\text{clr}}$ into an underspecified variant $\mathbf{x}_{\text{amb}}$ \yifan{that better reflects real-world medical query ambiguity} while preserving the original medical topic. 
The rewriting process applies the following transformation types that reflect major sources of real-world input ambiguity:
\begin{itemize}[leftmargin=*]
    \item \textbf{Context omission:} removing essential clinical details that are necessary for precise medical interpretation;
    \item \textbf{Semantic vagueness:} replacing specific descriptions with broader or less informative expressions;
    \item \textbf{Logical inconsistency:} introducing mild internal contradictions while keeping surface-level grammaticality.
\end{itemize}
\blue{The selection of transformation types is decided autonomously by the LLM based on its evaluation of question content. The full prompt template is provided in Appendix~\ref{app:prompt}, enabling the generation of ambiguous counterparts to be easily extended to other datasets and LLMs in practical use.} In Figure~\ref{fig:cvmedqa_example}, we show an example clear question paired with its rewritten ambiguous form.

\blue{To ensure the quality of the generated ambiguous questions, we conducted a human verification study that evaluated each rewritten variant along three dimensions: \textit{topic fidelity}, \textit{ambiguity validity}, and \textit{linguistic fluency}. Agreement rates exceeding 95\% confirmed that our method successfully introduced real-world ambiguity while preserving the original clinical context. Details of the validation protocol and quantitative results are reported in Appendix~\ref{app:human}.}

Ultimately, CV-MedBench provides contrastive question pairs $(\mathbf{x}_{\text{clr}},\, \mathbf{x}_{\text{amb}})$ that differ only in ambiguity level, which isolate the effect of input ambiguity from other linguistic or clinical factors. To the best of our knowledge, this is the first dataset designed to support systematic investigation of input ambiguity in Medical QA.


\begin{figure*}[t]
    \centering
    \includegraphics[width=\textwidth]{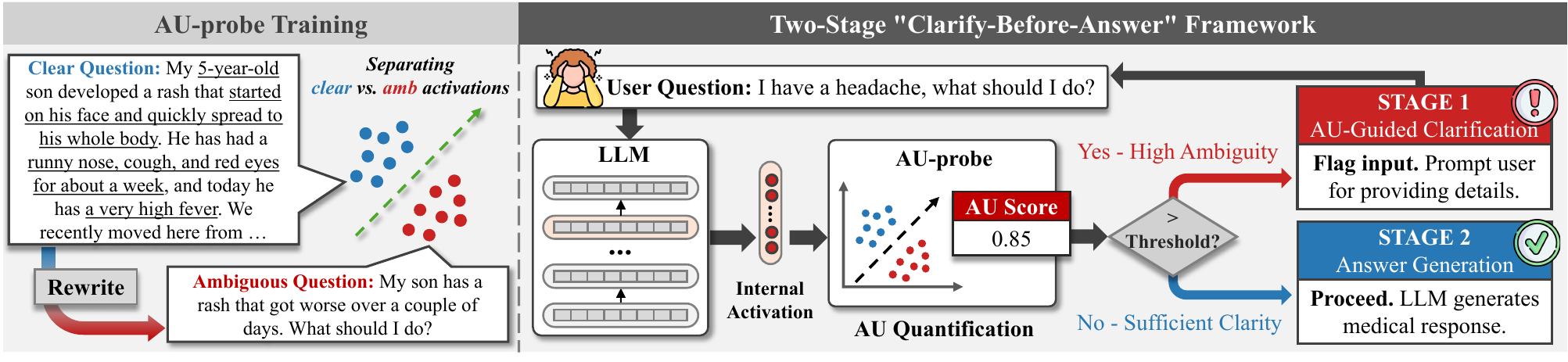}
    \vspace{-5mm}
    \caption{
        Overview of the proposed AU-guided ``Clarify-Before-Answer’’ framework. 
        Left: AU-Probe learns a linear separation between clear and ambiguous activations to predict AU.
        Right: the two-stage ``Clarify-Before-Answer’’ pipeline detects ambiguous inputs with AU score before answer generation, triggering clarification when needed or answering when the input is clear.
    }
    \label{fig:framework}
    \vspace{-2mm}
\end{figure*}

\section{Methodology}
\blue{In this section, we introduce our AU-guided ``Clarify-Before-Answer’’ framework for Medical QA, designed to mitigate safety risks caused by ambiguous inputs. Figure~\ref{fig:framework} provides an overview of the two-stage pipeline guided by the plug-in AU-Probe.}


\subsection{Definition and Problem Formulation}
\blue{Let $\mathbf{x}$ denote the input medical question and $\mathbf{y}$ denote its correct answer.
Formally, aleatoric uncertainty is defined as the entropy of the underlying answer distribution for $\mathbf{x}$:
\begin{equation}
U_{\text{aleatoric}}(\mathbf{x}) = \mathcal{H}\!\left(p(\mathbf{y}\,|\, \mathbf{x})\right),
\end{equation}
where $p(\mathbf{y}\,|\,\mathbf{x})$ denote the answer distribution. 
Aleatoric uncertainty characterizes the inherent stochasticity within the data-generating process and is irreducible even for a perfectly trained model. }
\yifan{In this work, we operationalize $U_{\text{aleatoric}}(\textbf{x})$ as the uncertainty arising from input ambiguity. In the Medical QA task, AU increases when underspecified symptom descriptions cause the answer distribution $p(\mathbf{y}\,|\,\mathbf{x})$ to disperse across multiple possible answers. Given a user input $\mathbf{x}$, we aim to estimate its aleatoric uncertainty $\hat{U}_{\text{aleatoric}}(\mathbf{x})$, which serves as an indicator of input ambiguity to safeguard the reliability of downstream answer generation.}

\subsection{Linear Encoding of Aleatoric Uncertainty}

Existing UQ methods often fail to provide robust AU estimation or incur high computational overhead (discussed in Sections~\ref{sec:intro} and ~\ref{sec:uq}). To overcome these limitations, we draw inspiration from the Linear Representation Hypothesis~\cite{park2023linear} \yifan{and quantify aleatoric uncertainty from intermediate LLM activations.}
Specifically, we utilize the LLM as a calibrated instrument to extract input ambiguity features, yielding an AU estimator that is both efficient and accurate.


\subsubsection{\textbf{Internal Activation Extraction}}

\blue{To investigate how aleatoric uncertainty is encoded within the LLM, we first extract hidden representations elicited by clear and ambiguous versions of the same medical question. }
For each contrastive pair $(\mathbf{x}_{\text{clr}},\, \mathbf{x}_{\text{amb}})$ \yifan{in CV-MedBench}, we independently feed both inputs into the LLM for \blue{medical QA} and record the residual-stream activations:
\begin{equation}
\big(\mathbf{a}^{(l)}(\mathbf{x}_{\text{clr}}), \mathbf{a}^{(l)}(\mathbf{x}_{\text{amb}})\big)
\quad\text{for } l = 1,\ldots,L,
\end{equation}
where $\mathbf{a}^{(l)}(\mathbf{x}) \in \mathbb{R}^{d}$ denotes the activation of the final prompt token at layer $l$, and $L$ is the total number of transformer layers. Layers are indexed from~1, and activations correspond to the outputs of transformer blocks.
\blue{We specifically extract the final prompt token's activation because it captures the model’s integrated semantic state for the entire input before answer generation, a representation known to be salient for high-level semantic signals of LLMs \cite{zou2022representation}.}


\subsubsection{\textbf{Linear Separability Analysis}}
\blue{Having obtained layer-wise activations for clear and ambiguous inputs, we next examine how these activations are organized within the LLM’s hidden space and whether they form distinguishable geometric patterns that reflect differences in aleatoric uncertainty.}

\begin{figure}[t]
    \centering
    \includegraphics[width=\linewidth]{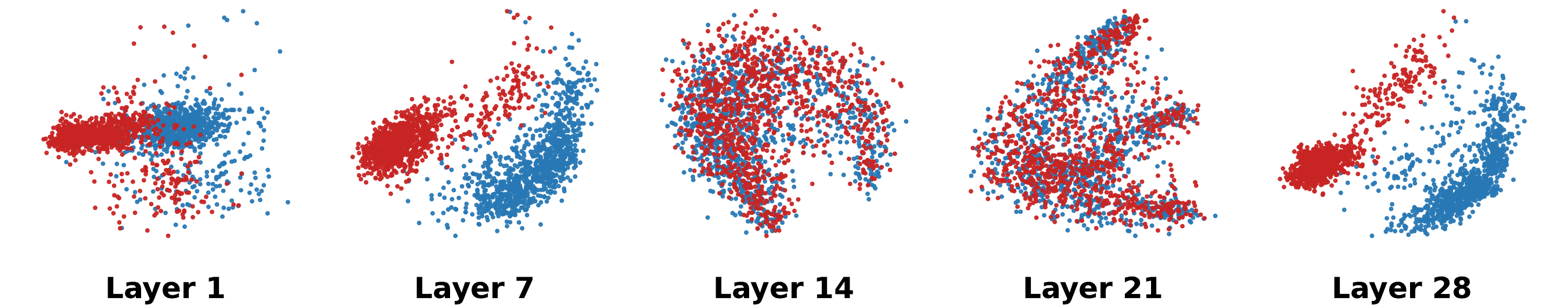}
    \caption{
        PCA visualization of activations for Qwen2.5-7B-Instruct \yifan{ demonstrate layer-dependent linear separability}. Blue: clear questions. Red: ambiguous questions.
    }
    \label{fig:pca_layers}
    \vspace{-5mm}
\end{figure}

\blue{In particular, we project the activations of contrastive question pairs into a two-dimensional subspace using PCA. Figure~\ref{fig:pca_layers} shows results for Qwen2.5-7B-Instruct, where clear and ambiguous questions form well-separated clusters in several layers of the network. This visual separation provides initial evidence that the internal representations are sensitive to input ambiguity. Critically, this linear separability suggests that the distinction between low-AU (clear) and high-AU (ambiguous) states can be defined by a simple hyperplane. Therefore, we posit that aleatoric uncertainty is encoded along a linear direction within the residual stream, a structure that enables quantification via a lightweight linear classifier.}

\blue{Due to space constraints, we only display a subset of layers and only one model. Similar separability patterns are observed across all models evaluated in our experiments (details in Section~\ref{sec:models}).}
\yifan{A quantitative analysis of layer-wise linear separability for all models is provided in Section~\ref{sec:layer}, which identifies the most separable layer for robust AU estimation.}

\subsection{Aleatoric Uncertainty Probe}

\blue{Building on the linear encoding of AU within hidden states, we introduce the Aleatoric Uncertainty Probe (AU-Probe), a lightweight plug-in module that estimates AU directly from internal activations of LLMs and requires no model fine-tuning or additional sampling, }\yifan{ enabling inference-time AU quantification for real-time medical QA systems.}

\subsubsection{\textbf{Probe Design and Training}}

For each transformer layer $l$, we train an independent linear probe to map the activation $\mathbf{a}^{(l)}(\mathbf{x}) \in \mathbb{R}^{d}$ to a scalar AU score:
\begin{equation}
\hat{u}^{(l)} = \sigma\!\left(\mathbf{w}^{(l)\top}\mathbf{a}^{(l)}(\mathbf{x}) + b^{(l)}\right),
\end{equation}
where $\sigma(\cdot)$ is the sigmoid function, and $\mathbf{w}^{(l)}, b^{(l)}$ are the layer-specific learnable parameters. The output $\hat{u}^{(l)}$ serves as the predicted aleatoric uncertainty score at layer $l$. Training is performed independently for all layers $l = 1,\ldots,L$. We optimize the probe parameters using the Binary Cross-Entropy (BCE) loss:
\begin{equation}
\mathcal{L}^{(l)}
= -\, u \log \hat{u}^{(l)} \;-\; (1-u) \log\!\left(1-\hat{u}^{(l)}\right),
\end{equation}
where clear questions are assigned the label $u = 0$ (low AU), whereas ambiguous questions are assigned the label $u = 1$ (high AU).


\blue{Notably, we adopt binary labels ($u \in \{0, 1\}$) instead of manually assigned continuous AU scores for two reasons.  
First, fine-grained aleatoric uncertainty annotation is inherently subjective and prone to high inter-annotator variability and label noise~\cite{wei2023fine,tanno2019learning}. Binary labels, conversely, provide a more stable training signal and substantially higher inter-annotator agreement in practice (Appendix~\ref{app:human}).
Second, despite binary supervision, the probe's sigmoid output produces continuous scores in $(0, 1)$, naturally allowing the probe to reflect the varying degrees of AU learned implicitly from the data distribution.
This design thus ensures stable supervision without sacrificing the ability to model graded aleatoric uncertainty.}

\subsubsection{\textbf{Layer Selection}}\label{sec:layer}

\begin{figure}[t]
    \centering
    \includegraphics[width=0.9\linewidth]{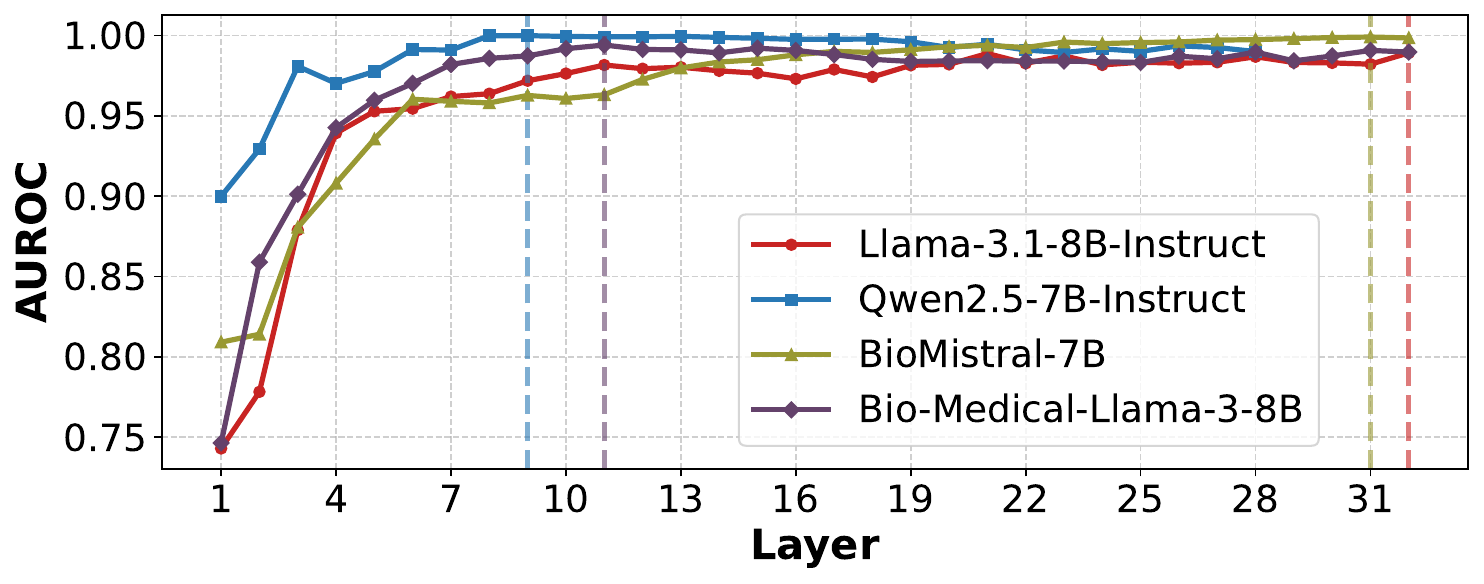}
    \vspace{-0.3cm}
    \caption{Layer-wise AUROC of the AU-Probe across LLMs.}
    \label{fig:layer_selection}
\vspace{-5mm}
\end{figure}

The strength of AU signals varies across layers. 
\yifan{Therefore, we identify the most informative layer by evaluating linear separability across all layers using a per-layer linear probe trained on the training set.}
Performance is measured by AUROC, where higher values indicate stronger linear separability between clear and ambiguous questions and thus a more informative AU signal.
As shown in Figure~\ref{fig:layer_selection}, the optimal layer differs across models.  
\yifan{On Llama-3.1-8B-Instruct and BioMistral-7B, the linear probe achieves peak AUROC in deeper layers (layers~28--32), whereas on Qwen2.5-7B-Instruct and Bio-Medical-Llama-3-8B, the strongest separability is observed in intermediate layers (layers~8--12). }



After selecting the most informative layer $l^{\ast}$, the linear probe trained on this layer serves as the AU-Probe. The final AU score for input question $\mathbf{x}$ is computed as:
\begin{equation}
\hat{U}_{\text{aleatoric}}(\mathbf{x})
= \sigma\!\left(\mathbf{w}^{(l^{\ast})\top}\mathbf{a}^{(l^{\ast})}(\mathbf{x}) + b^{(l^{\ast})}\right),
\end{equation}
yielding a continuous value in $(0,1)$. 
Higher AU scores indicate a greater likelihood of input ambiguity, while lower AU scores indicate clearer questions.
\blue{Operating solely on hidden activations of LLMs, the AU-Probe functions as a lightweight plug-in module that enables efficient and easily deployable AU quantification.}

\subsection{AU-Guided ``Clarify-Before-Answer’’ Medical QA Framework}

\blue{To address the safety risks posed by ambiguous user queries in medical QA, we integrate the AU-Probe into an AU-guided two-stage ``Clarify-Before-Answer" framework. As shown in Figure \ref{fig:framework}, our framework uses the AU score as an early signal to proactively detect ambiguous questions and ensure necessary clarification is obtained before generating a final response.}

\paragraph{\textbf{Stage 1: Ambiguity Detection and Clarification}} Given a user query $\mathbf{x}$, the AU-Probe computes $\hat{U}_{\text{aleatoric}}(\mathbf{x})$ directly from the hidden activation $\mathbf{a}^{(l^\ast)}(\mathbf{x})$, providing the model’s internal estimate of input ambiguity before answer generation. We trigger a clarification request by comparing \yifan{the AU score with }a threshold $\tau$: 
\[
\hat{U}_{\text{aleatoric}}(\mathbf{x}) > \tau \;\;\Longrightarrow\;\; 
\text{flag as ambiguous and request clarification.}
\]


\yifan{Empirically, we search thresholds in $\{0.1, 0.3, 0.5, 0.7, 0.9\}$ based on ambiguity–detection accuracy of AU-probe and find that $\tau = 0.5$ performs best across datasets and models (Fig.~\ref{fig:accuracy_threshold_robustness} in Appendix). The threshold remains tunable to balance safety and efficiency—for example, increasing $\tau$ reduces false alarms, whereas lowering it increases sensitivity to ambiguity.}




\paragraph{\textbf{Stage 2: Safe Answer Generation}}
\yifan{Once the query exhibits sufficiently low AU, either because the original input is already well-specified or because clarification has reduced the ambiguity, the system proceeds to answer generation:}
\[
\hat{U}_{\text{aleatoric}}(\mathbf{x}) \le \tau 
\;\;\Longrightarrow\;\;
\text{proceed to answer generation}.
\]
\blue{In this stage, the LLM executes answer generation, providing the final medical response to the well-specified query.}
\blue{This AU-guided "Clarify-Before-Answer" pipeline offers three advantages:}

\blue{\textbf{1. Plug-in and model-agnostic.}  
AU-Probe operates on hidden activations and requires no LLM fine-tuning, making the framework compatible and easily applied to any model used for medical QA.}

\blue{\textbf{2. Early safety intervention.}  
Ambiguity is detected before answer generation, preventing misinterpretation-induced errors and reducing the computational overhead of unnecessary decoding.}

\blue{\textbf{3. Improved reliability for real-world medical QA.}  
By prompting clarification when ambiguity is detected, the system delivers safer and more contextually grounded answers.}



\begin{table*}
\centering
\caption{Performance of UQ methods on medical QA datasets. Best AUROC (higher is better), ECE and Brier (lower is better) per model and dataset are in \textbf{bold}; second best are \underline{underlined}.}
\vspace{-2mm}
\label{tab:au_results}
\small
\setlength{\tabcolsep}{3.2pt}
\begin{tabular}{llccccccccc}
\toprule
\multirow{2}{*}{\textbf{Model}} & \multirow{2}{*}{\textbf{UQ Method}} & \multicolumn{3}{c}{\textbf{CV-MedQA}} & \multicolumn{3}{c}{\textbf{CV-MedMCQA}} & \multicolumn{3}{c}{\textbf{CV-MedExQA}} \\
\cmidrule(lr){3-5} \cmidrule(lr){6-8} \cmidrule(lr){9-11}
 &  & \textbf{AUROC $\uparrow$} & \textbf{ECE $\downarrow$} & \textbf{Brier $\downarrow$} & \textbf{AUROC $\uparrow$} & \textbf{ECE $\downarrow$} & \textbf{Brier $\downarrow$} & \textbf{AUROC $\uparrow$} & \textbf{ECE $\downarrow$} & \textbf{Brier $\downarrow$} \\
\midrule
\multirow{7}{*}{\textbf{Qwen2.5-7B-Instruct}} & MSP & 0.5573 & 0.4865 & 0.4850 & 0.5303 & 0.4816 & 0.4800 & 0.5299 & 0.4276 & 0.4359 \\
 & Mean Token Entropy & 0.5877 & 0.3029 & 0.3514 & 0.5421 & 0.2778 & 0.3433 & 0.5215 & 0.2902 & 0.3542 \\
 & Semantic Entropy & 0.5479 & 0.4825 & 0.4810 & 0.5233 & 0.4586 & 0.4594 & 0.5305 & 0.4395 & 0.4452 \\
 & SAR & 0.5546 & 0.3599 & 0.3882 & 0.5316 & 0.3164 & 0.3669 & 0.5376 & 0.3422 & 0.3819 \\
 & RAUQ & 0.5069 & 0.2487 & 0.3177 & 0.4938 & 0.2563 & 0.3256 & 0.5111 & \underline{0.2455} & 0.3147 \\
 & ASK4CONF & \underline{0.8281} & \underline{0.2429} & \underline{0.2529} & \underline{0.7190} & \underline{0.1837} & \underline{0.2540} & \underline{0.7056} & \textbf{0.2184} & \underline{0.2693} \\
 & \cellcolor{cyan!8} \textbf{AU-probe} & \cellcolor{cyan!8} \textbf{0.9998} & \cellcolor{cyan!8} \textbf{0.0515} & \cellcolor{cyan!8} \textbf{0.0109} & \cellcolor{cyan!8} \textbf{0.9999} & \cellcolor{cyan!8} \textbf{0.1035} & \cellcolor{cyan!8} \textbf{0.0207} & \cellcolor{cyan!8} \textbf{0.9987} & \cellcolor{cyan!8} 0.2488 & \cellcolor{cyan!8} \textbf{0.1277} \\
\midrule
\multirow{7}{*}{\textbf{Llama-3.1-8B-Instruct}} & MSP & 0.4971 & 0.4826 & 0.4820 & 0.5292 & 0.4797 & 0.4786 & 0.5514 & 0.4752 & 0.4737 \\
 & Mean Token Entropy & 0.3169 & 0.3084 & 0.3599 & 0.4900 & 0.2690 & 0.3354 & 0.5401 & 0.3317 & 0.3612 \\
 & Semantic Entropy & 0.5601 & 0.4412 & 0.4400 & 0.5560 & 0.4245 & 0.4260 & 0.5768 & 0.4369 & 0.4359 \\
 & SAR & 0.4865 & 0.2329 & 0.3186 & 0.5083 & 0.2005 & 0.3117 & 0.5364 & \underline{0.2150} & 0.3110 \\
 & RAUQ & 0.6405 & \underline{0.1595} & \underline{0.2540} & 0.5520 & \underline{0.1160} & \underline{0.2659} & 0.5600 & 0.2246 & \underline{0.3014} \\
 & ASK4CONF & \underline{0.7487} & 0.3840 & 0.3726 & \underline{0.6011} & 0.3665 & 0.3788 & \underline{0.6497} & 0.3886 & 0.3872 \\
 & \cellcolor{cyan!8} \textbf{AU-probe} & \cellcolor{cyan!8} \textbf{0.9891} & \cellcolor{cyan!8} \textbf{0.0411} & \cellcolor{cyan!8} \textbf{0.0424} & \cellcolor{cyan!8} \textbf{0.9690} & \cellcolor{cyan!8} \textbf{0.0984} & \cellcolor{cyan!8} \textbf{0.0794} & \cellcolor{cyan!8} \textbf{0.8867} & \cellcolor{cyan!8} \textbf{0.1839} & \cellcolor{cyan!8} \textbf{0.1793} \\
\midrule
\multirow{7}{*}{\textbf{Bio-Medical-Llama-3-8B}} & MSP & \underline{0.6341} & 0.4277 & 0.4280 & \underline{0.5833} & 0.4297 & 0.4331 & \underline{0.5905} & 0.4268 & 0.4287 \\
 & Mean Token Entropy & 0.4535 & 0.3168 & 0.3399 & 0.5645 & 0.1814 & 0.2876 & 0.5812 & 0.1619 & 0.2752 \\
 & Semantic Entropy & 0.6121 & 0.3329 & 0.3502 & 0.5308 & 0.2314 & 0.3117 & 0.5611 & 0.2657 & 0.3206 \\
 & SAR & 0.5317 & \textbf{0.0810} & \underline{0.2618} & 0.5441 & \textbf{0.0806} & \underline{0.2671} & 0.5572 & \textbf{0.0949} & \underline{0.2613} \\
 & RAUQ & 0.4993 & \underline{0.1624} & 0.2838 & 0.5168 & \underline{0.1293} & 0.2791 & 0.5371 & \underline{0.1385} & 0.2763 \\
 & ASK4CONF & 0.5458 & 0.3985 & 0.4005 & 0.5582 & 0.3845 & 0.3942 & 0.5546 & 0.3905 & 0.4000 \\
 & \cellcolor{cyan!8} \textbf{AU-probe} & \cellcolor{cyan!8} \textbf{0.9977} & \cellcolor{cyan!8} 0.1908 & \cellcolor{cyan!8} \textbf{0.0595} & \cellcolor{cyan!8} \textbf{0.9911} & \cellcolor{cyan!8} 0.2131 & \cellcolor{cyan!8} \textbf{0.0848} & \cellcolor{cyan!8} \textbf{0.9553} & \cellcolor{cyan!8} 0.2516 & \cellcolor{cyan!8} \textbf{0.1847} \\
\midrule
\multirow{7}{*}{\textbf{BioMistral-7B}} & MSP & 0.5853 & 0.4839 & 0.4847 & 0.4846 & 0.4585 & 0.4640 & \underline{0.6410} & 0.4644 & 0.4661 \\
 & Mean Token Entropy & 0.5221 & 0.3270 & 0.3596 & 0.4928 & 0.2599 & 0.3310 & 0.5981 & 0.2925 & 0.3301 \\
 & Semantic Entropy & 0.4453 & 0.2760 & 0.3493 & \underline{0.5361} & 0.2102 & 0.3074 & 0.6188 & 0.2168 & 0.2907 \\
 & SAR & 0.5490 & \underline{0.1244} & 0.2687 & 0.4960 & \underline{0.1195} & \underline{0.2739} & 0.6349 & \textbf{0.0741} & \underline{0.2434} \\
 & RAUQ & 0.5713 & 0.1913 & 0.2829 & 0.4844 & 0.1687 & 0.2891 & 0.6274 & \underline{0.1951} & 0.2778 \\
 & ASK4CONF & \underline{0.7554} & 0.1259 & \underline{0.2058} & 0.4356 & 0.3423 & 0.3910 & 0.5051 & 0.3365 & 0.3712 \\
 & \cellcolor{cyan!8} \textbf{AU-probe} & \cellcolor{cyan!8} \textbf{0.9997} & \cellcolor{cyan!8} \textbf{0.0379} & \cellcolor{cyan!8} \textbf{0.0102} & \cellcolor{cyan!8} \textbf{0.9958} & \cellcolor{cyan!8} \textbf{0.1063} & \cellcolor{cyan!8} \textbf{0.0388} & \cellcolor{cyan!8} \textbf{0.9780} & \cellcolor{cyan!8} 0.2282 & \cellcolor{cyan!8} \textbf{0.1551} \\
\bottomrule
\end{tabular}
\vspace{-2mm}
\end{table*}

\section{Experiments}
\subsection{Experimental Setup}
Full details of datasets, models, baselines, and evaluation protocols are provided in the Appendix~\ref{app:experimental_setup}.

\subsection{AU Quantification Performance}

Since AU stems from underspecified inputs, an effective AU quantification method should output higher scores to ambiguous questions and lower scores to clear ones, allowing the system to identify queries that pose ambiguity-related risk. 

As shown in Table~\ref{tab:au_results}, AU-Probe achieves near-perfect AUROC across all models, demonstrating its strong discriminative ability in 
in-distribution and out-of-distribution settings. 
These results verify that AU forms a linearly separable signal within the internal space of LLMs, and AU-Probe leverages this property effectively to achieve 
\yifan{reliable detection of input ambiguity across diverse test sets.}
\yifan{Regarding calibration, AU-Probe remains highly competitive. While methods like SAR sometimes yield lower ECE in OOD settings due to their tendency toward more conservative AU estimates, AU-Probe’s consistently superior Brier scores confirm that its sharper probability assignments can achieve a more accurate AU estimation.}

\yifan{Overall, AU-Probe benefits from direct access to internal hidden states. By operating on contextualized residual-stream activations, it bypasses noisy output-space signals, such as generation entropy or output likelihood utilized by methods like RAUQ. This allows AU-Probe to quantify AU more faithfully and robustly.}
\begin{table}[t]
\centering
\caption{QA Accuracy Improvement after AU-Guided Clarification (Threshold = 0.5). Improv. = +AU-probe over No Clar.}
\vspace{-5pt}
\small
\setlength{\tabcolsep}{4pt}
\begin{tabular}{lcccc}
\toprule
\textbf{Dataset} & \textbf{No Clar.} & \textbf{+ASK4CONF} & \textbf{+AU-probe} & \textbf{Improv.} \\
\midrule
\rowcolor{gray!10}\multicolumn{5}{c}{\textbf{Qwen2.5-7B}} \\
CV-MedQA & 0.4855 & 0.5169 & \textbf{0.5939} & \textbf{+10.84\%} \\
CV-MedMCQA & 0.5050 & 0.5130 & \textbf{0.5660} & \textbf{+6.10\%} \\
CV-MedExQA & 0.6415 & 0.6585 & \textbf{0.7309} & \textbf{+8.94\%} \\

\midrule
\rowcolor{gray!10}\multicolumn{5}{c}{\textbf{Llama-3.1-8B}} \\
CV-MedQA & 0.4815 & 0.5232 & \textbf{0.5719} & \textbf{+9.03\%} \\
CV-MedMCQA & 0.4810 & 0.4970 & \textbf{0.5410} & \textbf{+6.00\%} \\
CV-MedExQA & 0.6074 & 0.6170 & \textbf{0.6830} & \textbf{+7.55\%} \\

\midrule
\rowcolor{gray!10}\multicolumn{5}{c}{\textbf{Bio-Medical-Llama-3-8B}} \\
CV-MedQA & 0.6080 & 0.6866 & \textbf{0.7643} & \textbf{+15.63\%} \\
CV-MedMCQA & 0.5640 & 0.6340 & \textbf{0.7390} & \textbf{+17.50\%} \\
CV-MedExQA & 0.5926 & 0.6521 & \textbf{0.6957} & \textbf{+10.32\%} \\     

\midrule
\rowcolor{gray!10}\multicolumn{5}{c}{\textbf{BioMistral-7B}} \\
CV-MedQA & 0.3637 & 0.3841 & \textbf{0.4116} & \textbf{+4.79\%} \\
CV-MedMCQA & 0.3620 & 0.3880 & \textbf{0.4140} & \textbf{+5.20\%} \\
CV-MedExQA & 0.4596 & 0.5277 & \textbf{0.5787} & \textbf{+11.91\%} \\
\bottomrule
\end{tabular}
\label{tab:clarification-acc}
\vspace{-5mm}
\end{table}

\subsection{Effectiveness of AU-Guided Clarification}

\begin{figure*}[t]
    \centering
    \includegraphics[width=\textwidth]{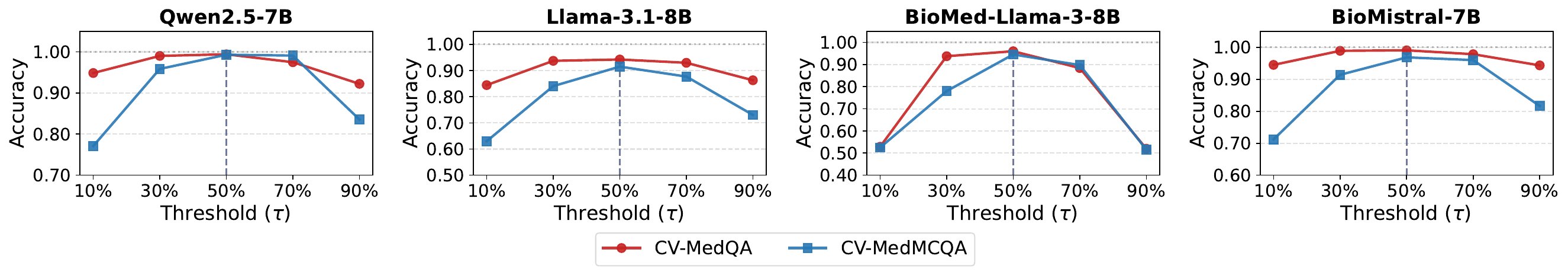} 
    \vspace{-6mm}
    \caption{AU-Probe ambiguity–detection accuracy across four LLMs under different ambiguity thresholds ($\tau$) on validation sets. Our threshold search shows that $\tau=0.5$ leads to the highest validation detection accuracy across different models.}
    \label{fig:accuracy_threshold_robustness}
\vspace{-3mm}
\end{figure*}

In this section, we evaluate whether AU-guided clarification driven by AU-Probe improves LLM answer accuracy and validate the two-stage “clarify-before-answer” framework. Experiments are conducted on the ambiguous versions of questions in the test set, representing realistic, underspecified user inputs in medical QA. \yifan{We report the QA accuracy of the base model (No Clarify.), ASK4CONF, which is the strongest baseline from our AUROC evaluation, and AU-Probe. The base model establishes a lower bound that struggles to resolve ambiguous queries.}
Both ASK4CONF and AU-Probe adopt $0.5$ as the threshold for ambiguity detection and triggering clarification. For ASK4CONF, this threshold is explicitly defined in its prompt (Figure~\ref{fig:ask4conf_prompt}); for AU-Probe, $0.5$ is evaluated as the optimal decision boundary (Figure~\ref{fig:accuracy_threshold_robustness}). 
In particular, if a question’s predicted AU exceeds $0.5$, it is treated as ambiguous and replaced with its clear counterpart to simulate user-provided clarification; otherwise, the original ambiguous question is retained. The QA accuracy of the LLM is computed on this clarified input stream.

\yifan{As shown} in Table~\ref{tab:clarification-acc}, 
AU-guided clarification consistently and significantly improves QA accuracy over the 
base model, confirming the efficacy of the two-stage paradigm: proactive ambiguity detection followed by targeted clarification substantially enhances response reliability in safety-critical medical QA.
Moreover, AU-Probe consistently outperforms ASK4CONF in guiding clarification decisions. This advantage stems from its stronger discriminative capability, achieved through a learned linear direction that more precisely separates clear and ambiguous representations in the activation space. As a result, AU-Probe triggers clarification more accurately, avoiding unnecessary interventions on clear inputs while reliably resolving genuinely ambiguous queries. 
These findings highlight the advantage of our AU-guided "clarify-before-answer" framework in enabling safer medical QA systems.

\begin{table}[t]
\centering
\caption{ID (underlined) and OOD AUROC results.}
\vspace{-3mm}
\small
\label{tab:ood}
\setlength{\tabcolsep}{6pt}
\begin{tabular}{lcccc}
\toprule
\textbf{Training} & \multicolumn{3}{c}{\textbf{Test sets}} \\
\cmidrule(lr){2-4}
\textbf{sets} & \textbf{CV-MedQA} & \textbf{CV-MedMCQA} & \textbf{CV-MedExQA} \\
\midrule
\rowcolor{gray!10}\multicolumn{4}{c}{\textbf{Qwen2.5-7B-Instruct}} \\
\textbf{CV-MedQA} & {\underline{0.9998}} & 0.9965 & 0.9996 \\
\textbf{CV-MedMCQA} & 0.9980 & {\underline{0.9998}} & 0.9996 \\
\midrule
\rowcolor{gray!10}\multicolumn{4}{c}{\textbf{Llama-3.1-8B-Instruct}} \\
\textbf{CV-MedQA} & {\underline{0.9959}} & 0.8944 & 0.8639 \\
\textbf{CV-MedMCQA} & 0.9785 & {\underline{0.9993}} & 0.9927 \\
\midrule
\rowcolor{gray!10}\multicolumn{4}{c}{\textbf{Bio-Medical-Llama-3-8B}} \\
\textbf{CV-MedQA} & {\underline{0.9962}} & 0.8742 & 0.8564 \\
\textbf{CV-MedMCQA} & 0.9860 & {\underline{0.9971}} & 0.9661 \\
\midrule
\rowcolor{gray!10}\multicolumn{4}{c}{\textbf{BioMistral-7B}} \\
\textbf{CV-MedQA} & {\underline{0.9995}} & 0.9262 & 0.9209 \\
\textbf{CV-MedMCQA} & 0.9380 & {\underline{0.9922}} & 0.9224 \\
\bottomrule
\end{tabular}
\vspace{-5mm}
\end{table}

\subsection{Robustness Analysis}
\subsubsection{\textbf{Out-of-Distribution Evaluation}}\label{sec:ood}


A key challenge for supervised uncertainty estimators is robustness under distribution shift. We evaluate AU-Probe in both in-distribution (ID) and out-of-distribution (OOD) settings. AU-Probe is trained separately on the training splits of CV-MedQA and CV-MedMCQA and evaluated on three held-out test splits (CV-MedQA, CV-MedMCQA, and CV-MedExQA). The CV-MedExQA dataset is used exclusively for inference as an OOD benchmark. Training and test splits remain strictly \textit{disjoint} to avoid leakage. Results are summarized in Table~\ref{tab:ood}.

In ID evaluation, AU-Probe achieves AUROC values close to~1.0 across all models, demonstrating that LLM hidden states encode a highly discriminative, linearly separable signal of AU, allowing the probe to accurately distinguish between clear and vague inputs. 
Under OOD evaluation, performance decreases slightly but remains strong, with AUROC consistently above 0.85. The result suggests that although the types of ambiguity differ across datasets, LLMs encode aleatoric uncertainty in a largely dataset-agnostic manner within their internal representations. As a result, AU-Probe generalizes effectively across datasets and maintains reliable AU quantification under distribution shift \yifan{as detailed in Section~\ref{sec:models}.} 

Among the tested models, Qwen2.5-7B-Instruct exhibits exceptional robustness, maintaining AUROC values near 1.0 in both ID and OOD settings. The results suggest that internal representations of Qwen2.5-7B-Instruct display a discriminative linear separation between clear and ambiguous questions, resulting in superior generalization and highlighting the probe’s effectiveness when trained on high-quality internal representations.

\subsubsection{\textbf{Low-Data Regime Performance}}\label{low-data}

\begin{figure}[t]
    \centering
    \includegraphics[width=\linewidth]{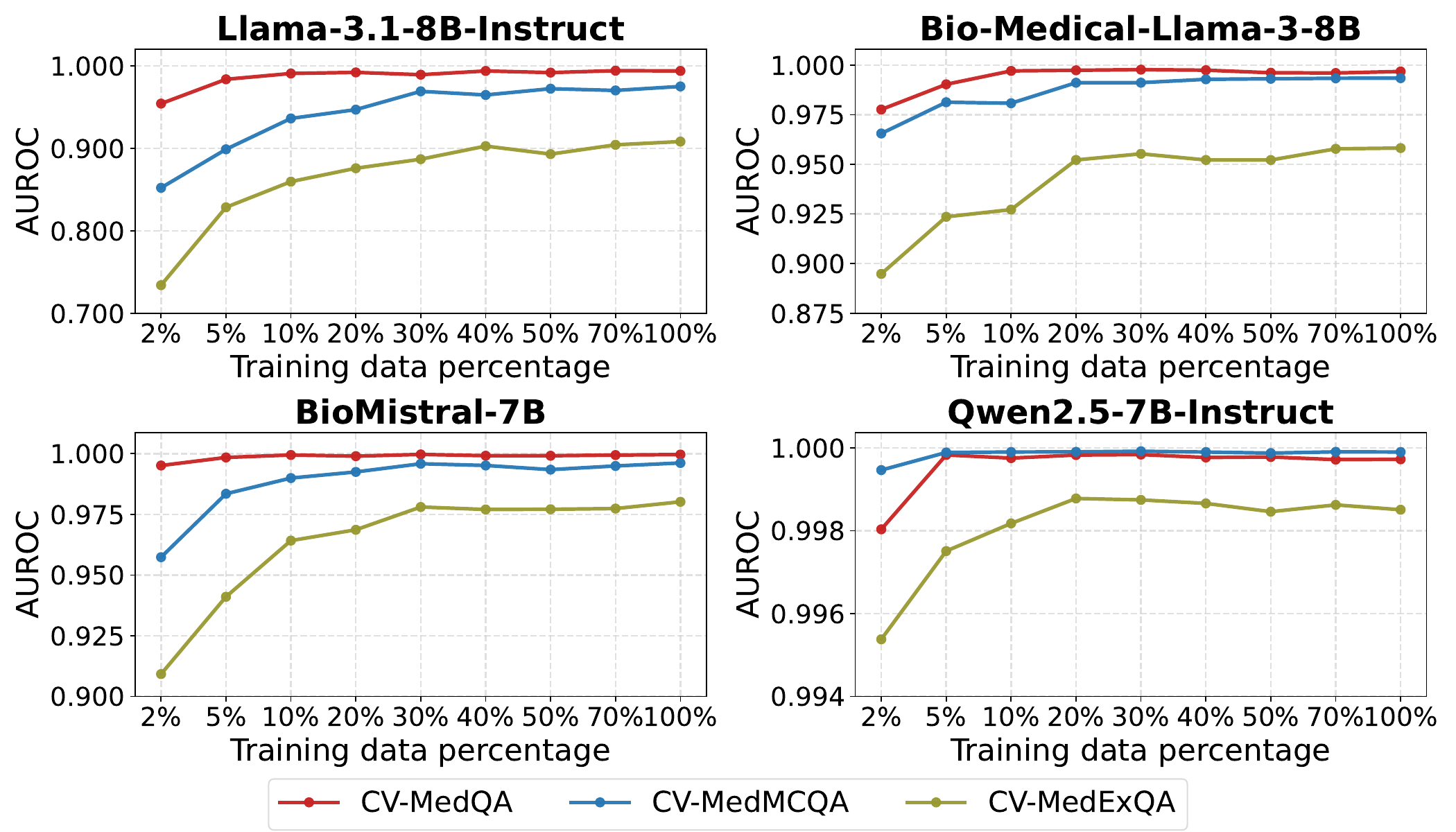}
    \vspace{-5mm}
    \caption{AU-Probe under different training sizes.}
    \label{fig:lowdata}
    \vspace{-5mm}
\end{figure}


We further evaluate the robustness of AU-Probe under limited labeled data. The goal is to determine how much supervision is needed for the probe to achieve stable AU quantification, providing practical guidance for low-resource settings.
Starting from the full CV-MedBench training splits (1,600 instances),  we progressively reduce the training size to 70\%, 50\%, 40\%, 30\%, 20\%, 10\%, 5\%, and 2\%. \yifan{Notably, the primary results reported in Tables~\ref{tab:au_results}, \ref{tab:clarification-acc}, \ref{tab:ood} utilize only the 30\% training size.} 

As shown in Figure~\ref{fig:lowdata}, AU-Probe exhibits rapid convergence and high data efficiency. On ID subsets (CV-MedQA and CV-MedMCQA), near-optimal AUROC is achieved with as little as 5–10\% of the training data for most models, and saturation occurs by 30\% at the latest. On the OOD CV-MedExQA subset, performance improves more gradually but stabilizes between 20\% and 40\% of the full training set. These results indicate that only a small fraction of training data is needed to achieve strong AU quantification.
Across models, Qwen2.5-7B-Instruct achieves peak performance with the fewest samples, suggesting that its hidden representations better capture the distinction between clear and ambiguous inputs. This finding is consistent with its strong robustness in the OOD evaluation.

Overall, using $30\%$ of the training data (960 samples) achieves performance nearly identical to full-data training, so we we use this reduced set for all AU-Probe models to improve computational efficiency. 
Remarkably, even with as little as $2\%$ of the data (64 samples), AU-Probe remains substantially stronger than the baseline methods in Table~\ref{tab:au_results}, indicating strong data efficiency and reliable performance in low-data settings with limited annotations.

\subsection{Computational Efficiency Analysis}

\begin{table}[t]
\centering
\caption{Per-sample inference latency (s) of UQ methods.}
\label{tab:uq_latency}
\vspace{-3mm}
\small
\begin{tabular}{lcccc}
\toprule
\textbf{Method} & \textbf{Qwen2.5} & \textbf{Llama3.1} & \textbf{BioMed-Llama} & \textbf{BioMistral} \\
\midrule
\textbf{MSP} & 1.142 & 1.143 & 0.977 & 1.210 \\
\textbf{MTE} & 1.185 & 1.148 & 0.982 & 1.214 \\
\textbf{SE} & 10.445 & 10.222 & 20.970 & 26.179 \\
\textbf{SAR} & 10.971 & 10.819 & 25.879 & 34.622 \\
\textbf{ASK4CONF} & 0.996 & 0.969 & 0.716 & 1.096 \\
\textbf{RAUQ} & 11.847 & 11.350 & 27.054 & 35.480 \\
\cellcolor{cyan!8}\textbf{AU-Probe} & \cellcolor{cyan!8}1.217 & \cellcolor{cyan!8}1.181 & \cellcolor{cyan!8}1.050 & \cellcolor{cyan!8}1.222 \\
\bottomrule
\end{tabular}
\vspace{-1mm}
\end{table}

For clinical deployment, an AU estimator must impose low computational overhead to preserve real-time interactivity. 
\yifan{In Table~\ref{tab:uq_latency}}, we benchmark all UQ methods on an NVIDIA A40 GPU, measuring per-sample \yifan{answer generation} time on the CV-MedQA test set.


\blue{Likelihood-based methods (MSP and MTE) and ASK4CONF are the fastest among baselines, with latencies around 0.9–1.2 seconds. However, despite their relatively low cost, these methods exhibit markedly weaker discrimination (Table~\ref{tab:au_results}), limiting their utility in safety-critical settings where missed ambiguity directly impacts patient risk.
Consistency-based methods (SE and SAR) and RAUQ incur substantially higher latency, ranging from 10 to over 35 seconds, depending on the model. Their cost is dominated by multi-sample generation (SE, SAR) or attention-head aggregation during decoding (RAUQ), making them impractical for real-time interactive clinical applications. By contrast, AU-Probe maintains low latency across all four models (1.05–1.22 seconds), comparable to the fastest baselines while avoiding the substantial overhead seen in multi-round methods. Combined with its significantly superior discrimination performance (Table~\ref{tab:au_results}), AU-Probe offers an advantageous balance of accuracy and efficiency, making it well-suited for timely ambiguity detection in medical QA workflows.}


\subsection{Case Study}

In this section, we present a real user case to illustrate how the AU-guided “clarify-before-answer’’ framework detects input ambiguity and enhances the safety of Medical QA.

\begin{figure}[t]
    \centering
    \includegraphics[width=\linewidth]{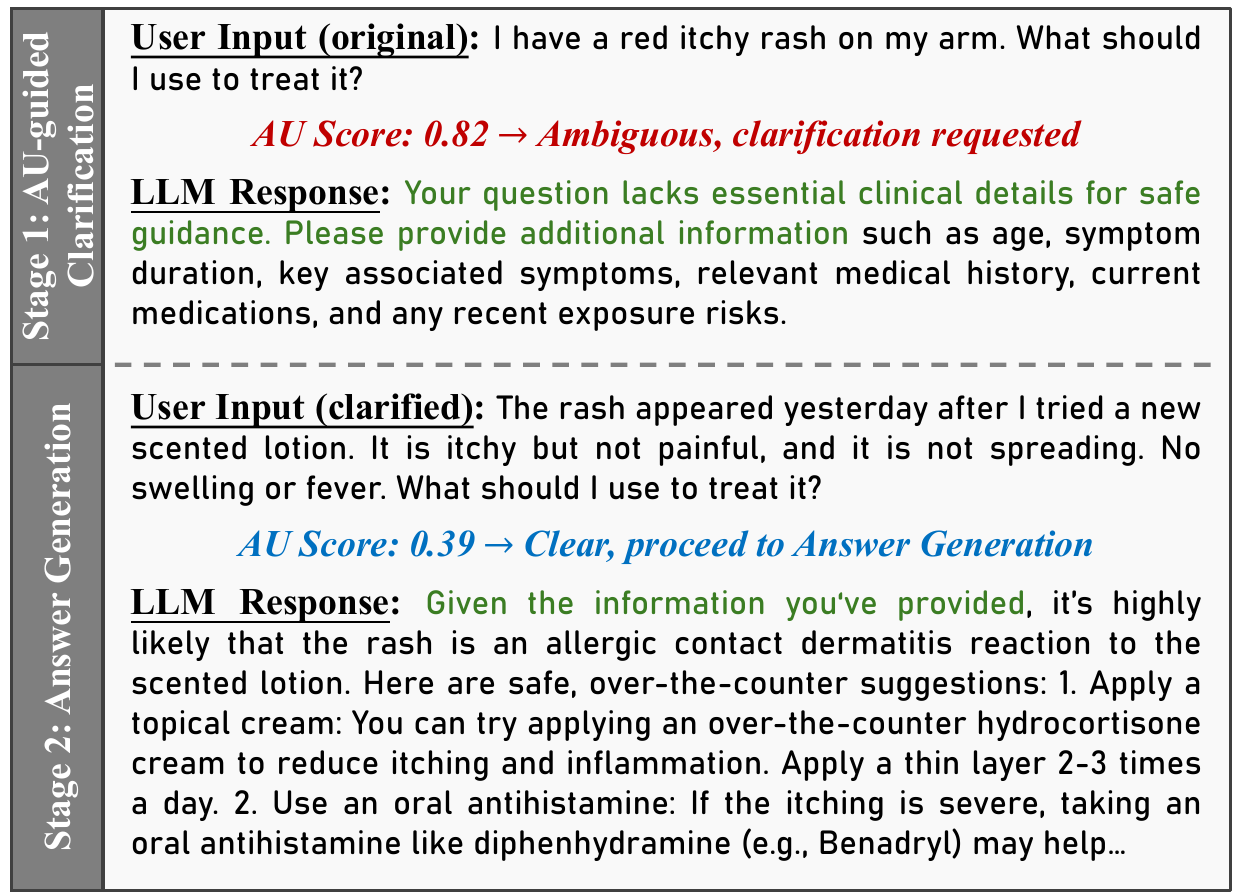}
    \vspace{-5mm}
    \caption{Case study of AU-guided clarification.}
    \label{fig:case}
    \vspace{-2mm}
\end{figure}

As shown in Figure~\ref{fig:case}, the user inputs the initial query (\textsf{“I have a red itchy rash on my arm. What should I use to treat it?”}) into Llama-3.1-8B-Instruct. This query lacks key diagnostic information, and the AU-Probe successfully assigns a high uncertainty score ($0.82$, above the $\tau=0.5$ threshold), triggering user clarification. The system requests essential clinical details before attempting to answer.
After the user provides the detailed information, the AU score decreases substantially ($0.39$), indicating that the input is now sufficiently well-specified. At this point, the model proceeds to the answer-generation stage and delivers a safe, context-aware recommendation consistent with standard care for contact dermatitis.

This case demonstrates the key benefit of the AU-guided “clarify-before-answer’’ framework: ambiguous queries are detected and intercepted successfully by the AU-probe before answer generation, clarification requests supply the missing context, and the final response becomes both safer and more clinically reliable.

\section{Limitations}

Our work presents two limitations that should be considered during the practical deployment of the AU-Probe.

First, the AU-Probe assesses ambiguity solely based on the linguistic formulation of the user query (i.e., how underspecified the prompt is). It does not detect factual inconsistencies or a semantic–reality mismatch where a clearly phrased question is based on incorrect underlying medical information from the user. Addressing this limitation would require incorporating external validation mechanisms (e.g., real-time physiological data), which is beyond the scope of our methods. Users should therefore interpret the uncertainty estimates as reflecting question ambiguity, not the objective factual completeness of the user's medical context.

Second, the AU-Probe requires white-box access to the internal hidden-state representations of the target LLM. This dependency restricts its application to models where internal activations are accessible and precludes its direct use with black-box commercial APIs that only expose the final output. Extending this methodology to black-box models remains an area for future research.

\section{Conclusion}

\yifan{This paper tackles the critical safety risks posed by input ambiguity in Medical QA by formulating it as quantifiable aleatoric uncertainty. Through a novel representation engineering perspective on uncertainty quantification, we introduce AU-Probe—a computationally lightweight module that estimates AU in a single pass—and integrate it into the first two-stage “clarify-before-answer” framework for proactive input ambiguity mitigation. Extensive evaluations demonstrate AU-Probe’s superior ambiguity quantification capability and the AU-guided QA framework’s substantial accuracy gains. By enhancing the reliability of LLMs in medical QA systems without compromising real-time inference efficiency, our work advances the broader vision of AI for social good.}

\begin{acks}
This research is supported in part by the National Science Foundation under Grant No. CNS-2427070,  IIS-2331069,  IIS-2202481, IIS-2130263, CNS-2131622. The views and conclusions contained in this document are those of the authors and should not be interpreted as representing the official policies, either expressed or implied, of the U.S. Government. The U.S. Government is authorized to reproduce and distribute reprints for Government purposes notwithstanding any copyright notation here on.
\end{acks}

\bibliographystyle{ACM-Reference-Format}
\balance
\bibliography{reference}

\begin{figure}[t]
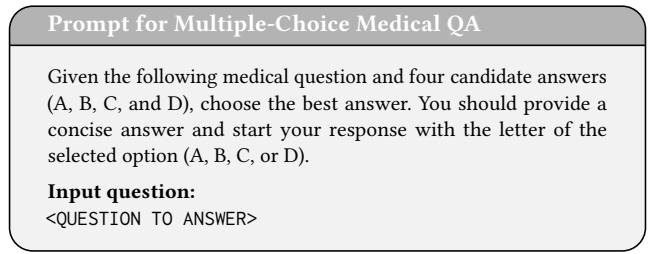

\centering
\begin{tcolorbox}[
    colback=gray!10,
    colframe=black,
    arc=3mm,
    boxrule=0.6pt,
    title={Prompt for Multiple-Choice Medical QA},
    coltitle=white,
    fonttitle=\bfseries,
    colbacktitle=gray!60,
    enhanced,
]
\small
Given the following medical question and four candidate answers (A, B, C, and D), choose the best answer.  
You should provide a concise answer and start your response with the letter of the selected option (A, B, C, or D).

\medskip
\textbf{Input question:}\\
\texttt{<QUESTION TO ANSWER>}

\end{tcolorbox}
\vspace{-5mm}
\caption{Prompt used for answering multiple-choice medical questions.}
\label{fig:mcq_prompt}
\end{figure}

\begin{figure}[t]
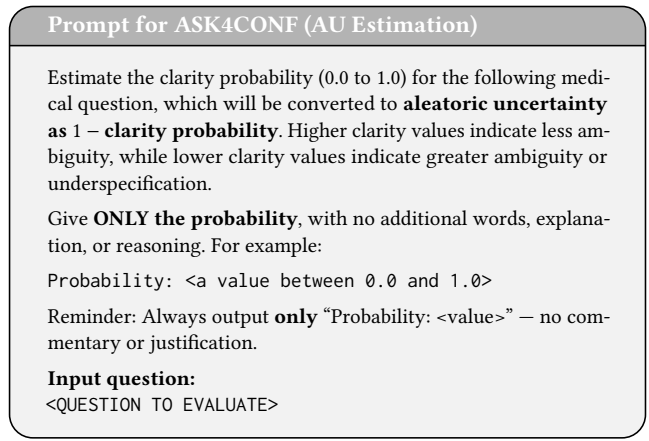

\centering
\begin{tcolorbox}[
    colback=gray!10,
    colframe=black,
    arc=3mm,
    boxrule=0.6pt,
    title={Prompt for ASK4CONF (AU Estimation)},
    coltitle=white,
    fonttitle=\bfseries,
    colbacktitle=gray!60,
    enhanced,
]
\small
Estimate the clarity probability (0.0 to 1.0) for the following medical question, which will be converted to  
\textbf{aleatoric uncertainty as $1 - \text{clarity probability}$}.  
Higher clarity values indicate less ambiguity, while lower clarity values indicate greater ambiguity or underspecification.

\medskip

Give \textbf{ONLY the probability}, with no additional words, explanation, or reasoning.  
For example:

\medskip
\texttt{Probability: <a value between 0.0 and 1.0>}

\medskip

Reminder: Always output \textbf{only} ``Probability: <value>'' — no commentary or justification.

\medskip
\textbf{Input question:}\\
\texttt{<QUESTION TO EVALUATE>}

\end{tcolorbox}
\vspace{-3mm}
\caption{Prompt used in the ASK4CONF setting for estimating aleatoric uncertainty ($1 - \text{clarity probability}$).}
\label{fig:ask4conf_prompt}
\end{figure}

\section{Appendix}

\subsection{Experimental Setup}\label{app:experimental_setup}
\subsubsection{\textbf{Dataset and Models}}\label{sec:models}

\blue{Our experiments are conducted on the Clear-to-Vague Medical QA Benchmark (CV-MedBench), constructed on questions from three publicly available multiple-choice medical QA datasets: MedQA \cite{medqa}, MedMCQA \cite{medmcqa}, and MedExQA \cite{medexqa}. These question sources span a wide range of clinical specialties, with MedQA and MedMCQA covering common subjects (e.g., pathology, physiology) and MedExQA focusing on underrepresented areas (e.g., Biomedical Engineering, Occupational Therapy). 
For the training split, we randomly sampled 800 questions each from the original MedQA and MedMCQA training splits. For testing, we adopt the original test splits of the three datasets, totaling 5,029 questions (1,273 from MedQA, 2,816 from MedMCQA, and 940 from MedExQA). 
All questions include both clear and ambiguous versions, and training and test splits are strictly disjoint.
Test sets from MedQA and MedMCQA are treated as in-distribution (ID), and MedExQA serves as out-of-distribution (OOD) due to its distinct specialty coverage and absence from training.
\textit{For clarity, we refer to the corresponding CV-MedBench test subsets as CV-MedQA, CV-MedMCQA, and CV-MedExQA according to their original sources.}}

\yifan{We conduct experiments using four open LLMs without fine-tuning, including both general-purpose instruction-tuned models (\textbf{Llama-3.1-8B-Instruct} and \textbf{Qwen2.5-7B-Instruct}) and specialized biomedical models (\textbf{Bio-Medical-Llama-3-8B} and \textbf{BioMistral-7B}) that are equipped with strong medical domain knowledge.} Our instruction prompts are in the Appendix~\ref{app:prompt}.

\blue{Notably, AU-Probe is only trained on 240 questions (each paired with clear and ambiguous versions) randomly sampled from the CV-MedBench training split, except for the low-data experiment in Section~\ref{sec:ood}, where all 800 questions are used to assess the amount of training data required. For AU quantification, each model uses the layer with the strongest AUROC signal identified in Section~\ref{sec:layer}:
Llama-3.1-8B-Instruct (layer 32), Qwen2.5-7B-Instruct (layer 9), BioMistral-7B (layer 30), and Bio-Medical-Llama-3-8B (layer 11).}

\subsubsection{\textbf{Baselines}}
We compare AU-Probe against a comprehensive set of UQ methods covering likelihood-based, consistency-based, LLM-as-a-judge, and internal-state-based methods:
\textbf{Likelihood-based} baselines estimate uncertainty directly from the model’s output distribution using a single forward pass, including Maximum Sequence Probability (MSP) and Max Token Entropy (MTE)~\cite{entropy}.  
\textbf{Consistency-based} methods assess uncertainty by measuring disagreement across multiple sampled generations.  
We evaluate Semantic Entropy (SE)~\cite{se} and SAR~\cite{sar}.  
\textbf{LLM-as-a-judge} methods rely on the model to explicitly score its own confidence under the default temperature.  
We adopt ASK4CONF~\cite{tian2023just}, using an adapted prompt that instructs the model to directly produce aleatoric uncertainty in the form of a clarity probability; the full prompt is provided in Appendix~\ref{app:prompt}.
\textbf{Internal-state-based} baselines infer uncertainty from hidden states.  We consider RAUQ~\cite{vazhentsev2025uncertainty} as a representative baseline. RAUQ differs fundamentally from our method: (1) it models total uncertainty rather than isolating aleatoric uncertainty, and (2) RAUQ combines attention-head features with output distributions and functions essentially as a likelihood-based method augmented by attention signals. In contrast, AU-Probe operates solely on residual-stream activations of the prompt token before any answer generation. This design decouples aleatoric uncertainty estimation from the generation process, enabling proactive detection of ambiguous inputs.

\subsubsection{\textbf{Evaluation Metrics}}
\blue{We evaluate AU quantification using both discrimination and calibration metrics. For discrimination, we report \textbf{AUROC}, which measures how effectively a method ranks ambiguous inputs above clear ones; higher values indicate a stronger discriminative ability. For calibration, we use \textbf{Expected Calibration Error (ECE)}~\cite{ece} to quantify the deviation between predicted AU scores and observed ambiguity frequencies, where lower values indicate better calibration. We additionally report the \textbf{Brier score}~\cite{brier}, which measures the mean squared error between predicted AU scores and ground-truth ambiguity, reflecting both calibration quality and prediction sharpness.}


\subsection{Prompt Templates}~\label{app:prompt}

We use the prompt in Figure~\ref{fig:dataset_prompt} to guide the LLM in rewriting clear medical questions into controlled ambiguous variants while preserving the original clinical topic. Figure~\ref{fig:mcq_prompt} shows the prompt used for answering multiple-choice questions, and Figure~\ref{fig:ask4conf_prompt} presents the ASK4CONF prompt for estimating aleatoric uncertainty from clarity probability.

\begin{figure*}[t]
\centering
\begin{tcolorbox}[
    colback=gray!10,
    colframe=black,
    arc=3mm,
    boxrule=0.6pt,
    title={Prompt for Generating Ambiguous Medical Questions},
    coltitle=white,
    fonttitle=\bfseries,
    colbacktitle=gray!60,
    enhanced,
]

\small
You are a medical data modifier. Your task is to rewrite clear and precise medical questions
to simulate real-world user queries that contain natural ambiguity (aleatoric uncertainty).

\medskip

You can apply one or more of the following uncertainty types, whichever are appropriate for the given question:

\textbf{1. Context omission} — remove important background information that makes the question less clear or underspecified.

Example 1: 

Original: ``A 60-year-old diabetic man with foot ulcer presents with fever and chills. What is the next best step in management?''  

Rewritten: ``A man with an infected wound on his foot feels unwell. What should the doctor do next?''  

Applied types: [``context omission'']

Example 2: 

Original: ``A 25-year-old woman with asthma develops sudden shortness of breath after taking aspirin.''  

Rewritten: ``A woman has trouble breathing after taking a medicine.''  

Applied types: [``context omission'']

\medskip

\textbf{2. Semantic vagueness} — wording becomes less precise or more general, while keeping the medical topic unchanged.

Example 1: 

Original: ``A 45-year-old man with pneumonia presents with productive cough and fever for 3 days. What antibiotic should be started?''  

Rewritten: ``A middle-aged man with a lung infection and fever for a few days needs treatment. What medicine should be given?''  

Applied types: [``semantic vagueness'']

Example 2: 

Original: ``A patient with rheumatoid arthritis is started on methotrexate. What lab tests should be monitored?''  

Rewritten: ``A patient on a new medication for joint pain needs regular tests. What kind of tests are usually done?''  

Applied types: [``semantic vagueness'']

\medskip

\textbf{3. Logical inconsistency} — introduce a small internal contradiction that may cause confusion but still reads naturally.

Example 1: 

Original: ``A newborn has jaundice with total bilirubin of 20 mg/dL. What is the next best step?''  

Rewritten: ``A newborn appears healthy but has a very high bilirubin level that is probably not serious. What should be done?''  

Applied types: [``logical inconsistency'']

Example 2: 

Original: ``A patient with chest pain and elevated troponin is diagnosed with myocardial infarction.''  

Rewritten: ``A patient with chest pain is told that their heart test is normal, but troponin is very high. What does this mean?''  

Applied types: [``logical inconsistency'']

\medskip

Do not change the main medical topic or intent of the question.  
If none of these types fit, return the original question unchanged.

\medskip

For each output, include:  

- The rewritten question  

- The applied uncertainty types (a list from [semantic vagueness, context omission, logical inconsistency])

\medskip
\textbf{Input question:} \\
\texttt{<QUESTION TO MODIFY>}
\end{tcolorbox}
\vspace{-3mm}
\caption{Prompt used to rewrite clear medical questions into ambiguous variants.}
\label{fig:dataset_prompt}
\end{figure*}

\subsection{Human Evaluation of Ambiguous Question Generation}~\label{app:human}

To validate the quality of the Clear-to-Vague Medical QA benchmark, we conducted a comprehensive human evaluation. The primary objective is to ensure the LLM successfully introduces ambiguity into the clear medical questions while maintaining the core clinical context and linguistic quality necessary for realistic scenarios. 
We evaluate each rewritten question along three binary dimensions (0 = fail, 1 = pass):
\begin{itemize}[leftmargin=*]
\item \textbf{Topic Fidelity (TF)}: whether the rewritten question preserves the same underlying medical topic or scenario as the original.
\item \textbf{Ambiguity Validity (AV)}: whether the rewritten question contains genuine, real-world ambiguity, consistent with context omission, semantic vagueness, or mild logical inconsistency. 
\item \textbf{Linguistic Fluency (LF)}: whether the rewritten question is grammatical, coherent, and natural in English. 
\end{itemize}
Two independent annotators reviewed each sample by comparing the original and rewritten questions and then assigning binary scores for all three evaluation dimensions. 

Table~\ref{tab:agreement_rate} reports the agreement rates between the two annotators across all evaluation dimensions. Agreement is high across all dimensions, with an overall agreement exceeding 95\%. The strong consensus validates that the rewritten variants successfully preserve clinical intent, introduce ambiguity, and maintain linguistic quality. Consequently, the Clear-to-Vague Medical QA dataset provide a solid resource for studying input ambiguity in medical QA.

\begin{table}[t]
\centering
\caption{Agreement rates across two human judges.}
\vspace{-3mm}
\label{tab:agreement_rate}
\small
\begin{tabular}{c|ccc}
\toprule
\multirow{2}{*}{\textbf{Dimension}} & \multicolumn{3}{c}{\textbf{Agreement Rate}} \\
 & \textbf{Human 1} & \textbf{Human 2} & \textbf{Combined} \\
\midrule
\textbf{TF} & 195/200 (97.5\%) & 195/200 (97.5\%) & 390/400 (97.5\%) \\ 
\textbf{AV} & 192/200 (96.0\%) & 192/200 (96.0\%) & 384/400 (96.0\%) \\ 
\textbf{LF} & 198/200 (99.0\%) & 198/200 (99.0\%) & 396/400 (99.0\%) \\ 
\bottomrule
\end{tabular}
\end{table}

\begin{figure}[t]
\centering
\begin{tcolorbox}[
    colback=gray!5,
    colframe=black!70,
    arc=2mm,
    boxrule=0.5pt,
    left=1.5mm,
    right=1.5mm,
    top=1mm,
    bottom=1mm
]
\small
\textbf{Clear Question:}
A 23-year-old pregnant woman at 22 weeks gestation presents with burning upon urination. She states it started 1 day ago and has been worsening despite drinking more water and taking cranberry extract. She otherwise feels well and is followed by a doctor for her pregnancy. Her temperature is 97.7\textdegree F (36.5\textdegree C), blood pressure is 122/77 mmHg, pulse is 80/min, respirations are 19/min, and oxygen saturation is 98\% on room air. Physical exam is notable for an absence of costovertebral angle tenderness and a gravid uterus. Which of the following is the best treatment for this patient?

\medskip
\textbf{Ambiguous Question:}
A pregnant woman feels pain while urinating and it's getting worse even after trying home remedies. She seems generally okay and has been seeing a doctor for her pregnancy. What is the best way to help her?
\end{tcolorbox}
\vspace{-0.5cm}
\caption{Example question from CV-MedBench.}
\label{fig:cvmedqa_example}
\end{figure}

\end{document}